\documentclass[sigconf]{acmart}
\usepackage{subcaption}
\usepackage{balance}
\DeclareMathOperator*{\argmaxA}{arg\,max} 

\AtBeginDocument{%
  \providecommand\BibTeX{{%
    \normalfont B\kern-0.5em{\scshape i\kern-0.25em b}\kern-0.8em\TeX}}}

\setcopyright{acmcopyright}
\copyrightyear{2021}
\acmYear{2021}
\setcopyright{acmcopyright}\acmConference[MM '21]{Proceedings of the 29th ACM International Conference on Multimedia}{October 20--24, 2021}{Virtual Event, China}
\acmBooktitle{Proceedings of the 29th ACM International Conference on Multimedia (MM '21), October 20--24, 2021, Virtual Event, China}
\acmPrice{15.00}
\acmDOI{10.1145/3474085.3475220}
\acmISBN{978-1-4503-8651-7/21/10}

\begin{document}
\fancyhead{}

\title{SimulLR: Simultaneous Lip Reading Transducer \\
with Attention-Guided Adaptive Memory}


\author{Zhijie Lin}
\authornote{Both authors contributed equally to this research.}
\affiliation{%
  \institution{Zhejiang University}
  \city{}
  \country{}}
\email{linzhijie@zju.edu.cn}

\author{Zhou Zhao}
\authornote{Zhou Zhao is the corresponding author.}
\affiliation{%
  \institution{Zhejiang University}
  \city{}
  \country{}
}
\email{zhaozhou@zju.edu.cn}

\author{Haoyuan Li}
\authornotemark[1]
\affiliation{%
  \institution{Zhejiang University}
  \city{}
  \country{}
}
\email{lihaoyuan@zju.edu.cn}

\author{Jinglin Liu}
\affiliation{%
  \institution{Zhejiang University}
  \city{}
  \country{}
}
\email{jinglinliu@zju.edu.cn}

\author{Meng Zhang}
\affiliation{%
  \institution{Huawei Noah's Ark Lab}
  \city{}
  \country{}
}
\email{zhangmeng92@huawei.com}

\author{Xingshan Zeng}
\affiliation{%
  \institution{Huawei Noah's Ark Lab}
  \city{}
  \country{}
}
\email{zxshamson@gmail.com}

\author{Xiaofei He}
\affiliation{%
  \institution{Zhejiang University}
  \city{}
  \country{}
}
\email{xiaofei_h@qq.com}

\renewcommand{\shortauthors}{Trovato and Tobin, et al.}

\begin{abstract}
  Lip reading, aiming to recognize spoken sentences according to the given video of lip movements without 
  relying on the audio stream, has attracted great interest due to its application in many scenarios.
  Although prior works that explore lip reading have obtained salient achievements, 
  they are all trained in a non-simultaneous manner where the predictions are generated requiring
  access to the full video.
  To breakthrough this constraint, we study the task of simultaneous lip reading and devise SimulLR,
  a simultaneous lip Reading transducer with attention-guided adaptive memory from three aspects:
  (1) To address the challenge of monotonic alignments while considering the syntactic structure of the generated sentences under simultaneous setting,
  we build a transducer-based model and design several effective training strategies including CTC pre-training, model warm-up and curriculum learning to promote the training
  of the lip reading transducer.
  (2) To learn better spatio-temporal representations for simultaneous encoder, we construct a truncated 3D convolution and time-restricted
  self-attention layer to perform the frame-to-frame interaction within a video segment containing fixed number of frames.
  (3) The history information is always limited due to the storage in real-time scenarios, 
  especially for massive video data. Therefore, we devise a novel attention-guided adaptive memory to organize semantic information of history segments and enhance the
  visual representations with acceptable computation-aware latency.
  The experiments show that the SimulLR achieves the translation speedup 9.10$\times$ compared with the state-of-the-art
  non-simultaneous methods, and also obtains competitive results, which indicates the effectiveness of our proposed methods.
\end{abstract}





\maketitle

\section{Introduction}
Lip reading, aiming to recognize spoken sentences according to the given video of lip movements without 
relying on the audio stream, has attracted great interest
~\cite{chung2017lip,martinez2020lipreading,assael2016lipnet,stafylakis2017combining,petridis2018audio,afouras2018deep,xu2018lcanet,zhang2019spatio} 
due to the application in many scenarios
including dictating instructions in public areas or a noisy environment, and providing help for
hard-of-hearing people. 
It remains a challenging task even for excellent lip readers~\cite{assael2016lipnet}.

\begin{figure}[t]
  \includegraphics[width=1.0\linewidth]{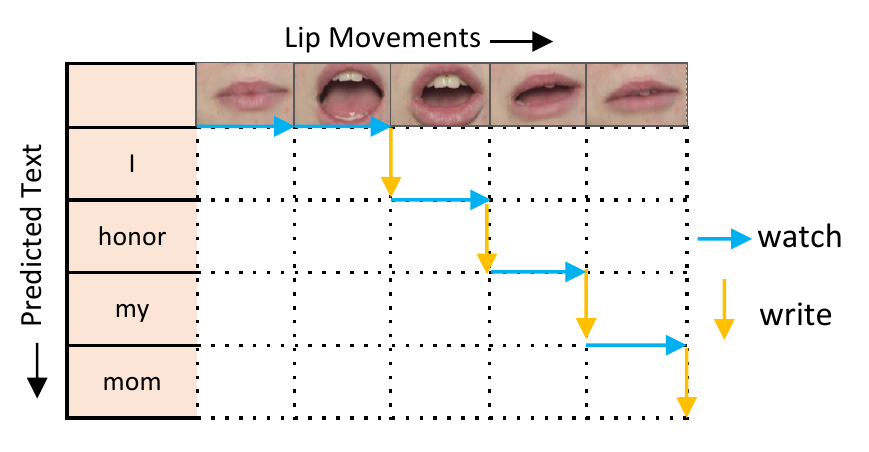}
  \caption{
        The frame-synchronized simultaneous decoding of the proposed lip reading transducer.
        At each time step, an empty transfer~(watch) is allowed to read next video segment or 
        a context-aware token can be generated~(write).
  }
  \label{fig:example}
\end{figure}

Although prior works that explore lip reading 
have obtained salient achievements, they are all trained in a non-simultaneous manner where the predictions are generated requiring
access to the full video. 
Therefore, simultaneous lip reading, where a video segment containing fix number of frames is processed
while spoken sentence is generated concurrently, 
is a more difficult but necessary extension for real-time understanding~(e.g. live video streaming).
Due to the low latency of simultaneous decoding, simultaneous lip reading are able to deal with massive video data~(e.g. long films)
without ``watching'' the entire video first.
In this paper, we study the task of simultaneous lip reading that recognizes sentences based on partial input.
However, it is very challenging to decode simultaneously for vision-text cross-modal translation in following aspects:

Firstly, for simultaneous decoding, 
the model is required to learn the monotonic alignments
between video segments and target tokens, and pick a suitable moment that achieves a good trade-off between latency and accuracy
to predict the next token. 
Due to the significant discrepancy of length of same token in different videos, it is difficult to estimate the duration of tokens
and learn such monotonic alignments. 
Prior autoregressive methods~\cite{chung2017lip,afouras2018deep,zhang2019spatio,martinez2020lipreading,zhao2020hearing}
leverage the semantic information of entire videos and work in a word-synchronized mode without considering monotonic alignments, 
making it non-simultaneous in nature.
A naive method is to scale the CTC-based model~\cite{assael2016lipnet,stafylakis2017combining,petridis2018audio,xu2018lcanet,chen2020duallip}
to simultaneous decoding by limiting each frame to see only its previous frames. 
However, the target sentences always show a strong correlation across time~\cite{liu2020fastlr}, but the CTC-based model
generate different tokens conditionally independent of each other, ignoring the syntactic information.
In our paper, inspired by neural transducer~\cite{graves2012sequence,jaitly2016online}, we devise a lip reading transducer that generates
tokens in a frame-synchronized mode where an empty transfer is allowed
to read next video segment at each time step~(See Figure~\ref{fig:example}), and also considers the syntactic structure of the generated sentences.
With the reading of video segments, the tokens generate frame-by-frame and then are merged to the ultimate predictions. 
We also design several effective training strategies including CTC pre-training, model warm-up and curriculum learning to promote the training
of lip reading transducer.

Secondly, to learn better spatio-temporal representations for cross-modal decoding, prior non-simultaneous methods~\cite{zhang2019spatio,liu2020fastlr}
employ multiple 3D convolution and self-attention layers in the visual encoder, which cannot be transferred to our simultaneous model due to their 
expanding receptive field on the whole video.
To obtain a better simultaneous encoder and reduce the gap between our method and non-simultaneous methods, we construct a truncated 3D convolution 
for spatio-temporal representations learning and time-restricted self-attention layer 
to perform the frame-to-frame interaction among available video segments.

Thirdly, in real scenarios, the storage is always limited by the extremely long input sequence~(e.g. massive video data).
Therefore, for simultaneous decoding, history segments may also be unavailable, making it more difficult to predict a new token with limited visual context.
To achieve a good storge-accuracy trade-off, inspired by memory networks~\cite{graves2014neural,graves2016hybrid},
we devise a novel attention-guided adaptive memory to organize semantic information of history segments and enhance the
visual representations using limited context.
Also, given the memory, the computation of the commonly-used self-attention mechanism is no longer conducted over all the history segments,
which reduces the computation-aware latency for simultaneous decoding~\cite{ma2020simulmt}.
Specially, the attention-guided memory is constructed to absorb new segments by momentum update and discard obsolete
features using the least frequently used~(LFU) algorithm guided by attention scores.
Based on the proposed adaptive memory, the simultaneous model incorporates both global context and adjacent semantic information
with acceptable computation-aware latency.

In summary, we study the task of simultaneous lip reading with limited history
and devise a vision-text cross-modal transducer SimulLR and devise
several effective training strategies to promote the performance.
For the simultaneous encoder, we construct a truncated 3D convolution and time-restricted self-attention layer
to learn better spatio-temporal representations for video segments. 
Further, considering the limited storage and computational cost,
we further devise a novel attention-guided adaptive memory to organize semantic information of history segments
for simultaneous decoding with acceptable computation-aware latency.

The experiments show that the SimulLR achieves the translation speedup 9.10$\times$ compared with the state-of-the-art
non-simultaneous methods, and also obtains competitive results, which indicates the effectiveness of our proposed methods.

\section{Related Works}
\subsection{Lip Reading}
Lip reading aims to recognize spoken sentences according to the given video of lip movements without 
relying on the audio stream. 
Early works focus on single word classification~\cite{chung2016lip,wand2016lipreading} 
and then switched to full sentences prediction~\cite{assael2016lipnet,stafylakis2017combining,chung2017lip,petridis2018audio,afouras2018deep,xu2018lcanet,zhang2019spatio}.
These works mainly study lip reading in a non-simultaneous manner with CTC-based model~\cite{assael2016lipnet,stafylakis2017combining,petridis2018audio,xu2018lcanet,chen2020duallip}
and autoregressive model~\cite{chung2017lip,afouras2018deep,zhang2019spatio,martinez2020lipreading,zhao2020hearing}.
Among them, LipNet~\cite{assael2016lipnet} takes advantage of spatio-temporal convolutional features and context modeling of RNNs.
Chen et al.~\cite{chen2020duallip} design a system that leverages the task duality of lip reading and lip generation to improve both tasks.
Afouras et al.~\cite{afouras2018deep} first introduce Transformer self-attention architecture into lip reading.
Zhao et al.~\cite{zhao2020hearing} enhance the training of lip reading model by distilling multi-granularity knowledge 
from speech recognition.
Besides, instead of CTC decoder, Liu et al.~\cite{liu2020fastlr} further study non-autoregressive lip reading by leveraging integrate-and-fire module to 
estimate the length of output sequence and alleviate the problem of time correlation.

However, these methods explore lip reading in a non-simultaneous manner, where the sentence prediction 
relies on the entire video of talking face during inference.
In this paper, we further study the task of simultaneous lip reading that recognizes sentences based on partial input, 
which owns more application scenarios.

\subsection{Simultaneous Decoding}
Due to lower latency and broader scenarios, simultaneous decoding has attracted a lot interest in many fields such as
neural machine translation~(NMT)~\cite{gu2016learning,grissom2014don}, 
automatic speech recognition~(ASR)~\cite{rao2017exploring,zhang2020unified,li2020towards,sainath2019two}, 
speech to text translation~\cite{oda2014optimizing,dalvi2018incremental,ren2020simulspeech,chen2020developing},
speech to speech translation~\cite{sudoh2020simultaneous} and so on.
In real-time scenarios, the simultaneous decoding aims to generate the predictions based on the given partial input
instead of the whole sequence, and the history context could be limited due to the rapid increase in the length of input.
Some widely used approaches for simultaneous decoding includes 
reinforcement learning~(RL)~\cite{gu2016learning,grissom2014don},
connectionist temporal classification~(CTC)~\cite{amodei2016deep}, 
transducers~\cite{rao2017exploring,sainath2019two,chen2020developing} 
and attention-based encoder-decoder~\cite{ren2020simulspeech,moritz2019triggered}.
Among them, at each time step, transducers generate the next target token, or an empty transfer to read next source input.

In this paper, we concentrate on vision-text cross-modal simultaneous decoding 
and propose a novel lip reading transducer with an adaptive memory
where the history frames are limited.

\begin{figure*}[t]
  \includegraphics[width=1.0\linewidth]{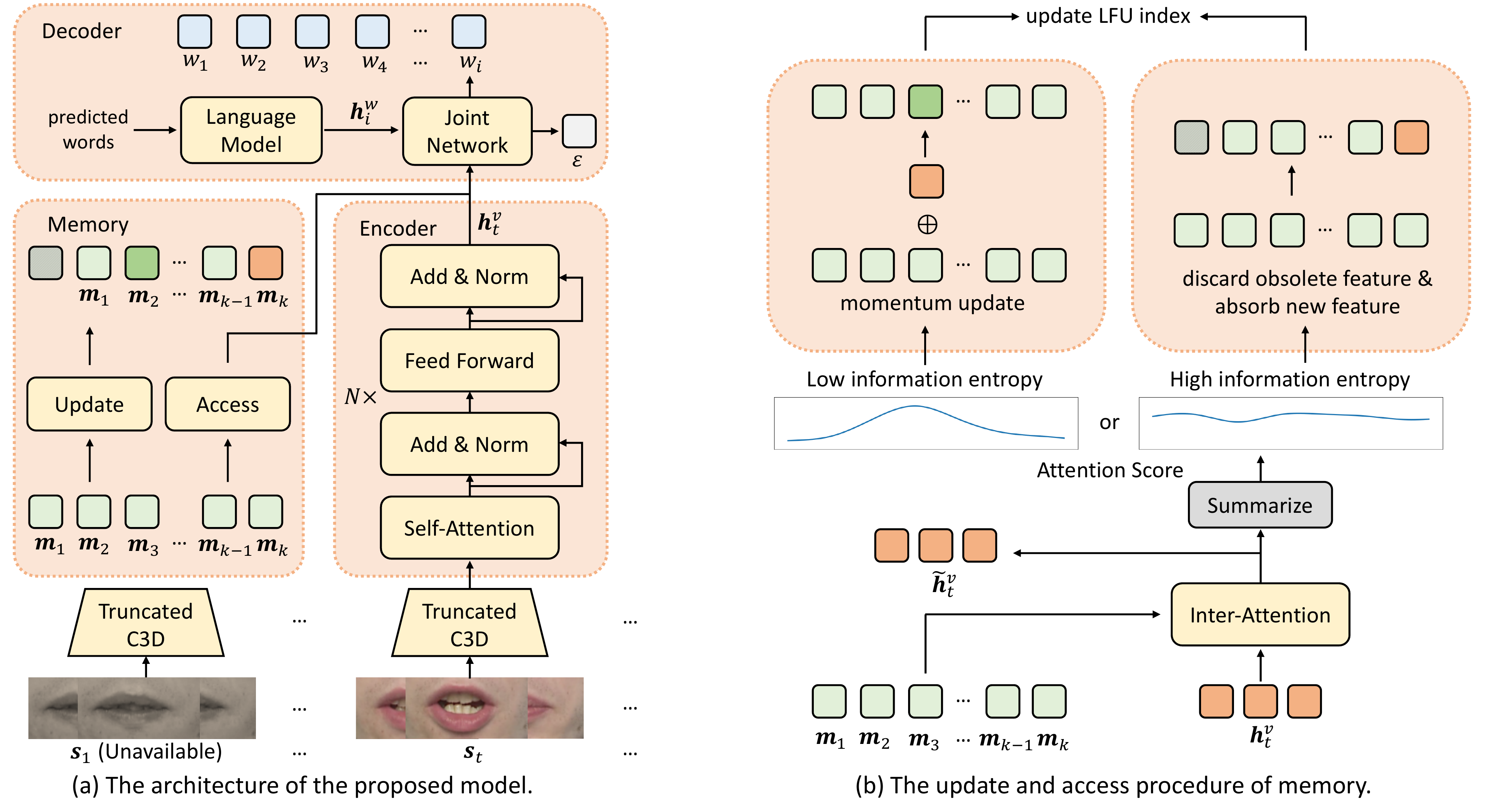}
  \caption{
        (a) The overall framework of SimulLR: a truncated 3D spatio-temporal convolutional network to extract the visual features, a transformer-based sequence
        encoder, a transducer-based cross-modal decoder for language modeling and token prediction,
        an attention-guided adaptive memory to organize semantic information of history segments and enhance the
        visual representations.
        (b) The update and access of memory: absorb new segments by momentum update and discard obsolete
        features using the least frequently used~(LFU) algorithm guided by attention scores.
  }
  \label{fig:framework}
\end{figure*}

\subsection{Memory}
Memory module introduces external memory to store the past context and absorb new information, which is proposed to
improve the learning capability and boost the performance.
The neural turing machine~(NTM)~\cite{graves2014neural} and differentiable neural computer~(DNC)~\cite{graves2016hybrid}
are the typical memory for memorization and reasoning.
For few-shot learning, memory module mainly stores the information contained in the support set~\cite{munkhdalai2017meta,vinyals2016matching}
and attempts to learn the common access mechanism across tasks.
Memory module has also been incorporated into generative models~\cite{bornschein2017variational,li2016learning}
and sequence modeling~\cite{le2018variational} that conditions on the global contextual information provided in external
memory. The recurrent neural networks such as GRU~\cite{chung2014empirical} are also commonly-used differentiable memory
module for sequence modeling, although they still suffer from gradual forgetting of early contents after memorizing long sequences.

In this paper, considering the limited storage and computational cost, 
we devise a novel attention-guided adaptive memory module to compress the history semantic
information and absorb upcoming video segments.

\section{Problem Formulation}
In this section, we first introduce the problem formulation of simultaneous lip reading.
Given a sequence of video segments ${\boldsymbol s} = \{{\boldsymbol s}_{1},{\boldsymbol s}_{2},...,{\boldsymbol s}_{n}\}$ 
without the audio stream, lip reading aims to predict the words sequence 
${w} = \{{w}_{1},{w}_{2},...,{w}_{u}\}$ that the lip is speaking, where ${\boldsymbol s}_{t}$ is the $t$-th video segments
containing several frames, $n$ is the number of video segments, $n_f$ is the number of frames in a segment,
${w}_{i}$ is the $i$-th token and $u$ is the length of target sequence.
Under the simultaneous setting, the lip reading model is required to generate the $i$-th token ${w}_{i}$ with only partial input
${\boldsymbol s_{p_{i}}} = \{{\boldsymbol s}_{1},{\boldsymbol s}_{2},...,{\boldsymbol s}_{n(w_i)}\}$, where $n(w_i)$ is
the number of segments needed to predict the $i$-th token ${w}_{i}$ and $n(w_i) >= n(w_{i-1})$ for monotonic alignments.
Also, in our paper, only the adjacent segments are available due to the limited storage, making the partial input
${\boldsymbol s_{p_{i}}} = \{{\boldsymbol s}_{n(w_i)-a+1},{\boldsymbol s}_{n(w_i)-a+2},...,{\boldsymbol s}_{n(w_i)}\}$,
where $a$ is the number of available segments for the $i$-th token prediction.
For simultaneous lip reading model, the monotonic alignments to predict the target sequence $w$ are not given explicitly,
which means that the decoding segment path ${d} = \{\boldsymbol s_{p_{1}},{\boldsymbol s_{p_{2}}},...,{\boldsymbol s_{p_{u}}}\}$
is not unique. Therefore, the optimize object can be computed as follows:
\begin{eqnarray}
  {P}(w|{\boldsymbol s}) = \sum_{{d \in \phi(w)}}\prod_{i=1}^{u}{P}(w_i|{\boldsymbol s_{p_{i}}})
\end{eqnarray}
where ${P}(w|{\boldsymbol s})$ is the probability of generating the target sequence $w$, which is the sum
over all possible decoding segment paths $d \in \phi(w)$.

\section{Approaches}
In this section, we describe the SimulLR approach thoroughly.
As shown in Figure~\ref{fig:framework}(a), the proposed model is composed of a truncated 
3D spatio-temporal convolutional network to extract the visual features, a transformer-based sequence
encoder, a transducer-based cross-modal decoder for language modeling and token prediction,
an attention-guided adaptive memory to organize semantic information of history segments and enhance the
visual representations with acceptable computation-aware latency.

We also design several effective training strategies including CTC pre-training, model warm-up and curriculum learning to promote the training
of the lip reading transducer.
The details of our method are described in the following subsections.

\subsection{Visual Encoder}
\textbf{Truncated C3D.}
To learn better spatio-temporal representations for cross-modal decoding, prior non-simultaneous methods~\cite{zhang2019spatio,liu2020fastlr}
employ multiple 3D convolution in the visual encoder, which cannot be transferred to our simultaneous model directly 
due to their expanding receptive field on the whole video. 
To address this challenge, in our paper, we truncate the 3D convolutional network in the temporal dimension and perform
spatio-temporal convolution only within one single segment ${\boldsymbol s}_{t}$, as shown in Figure~\ref{fig:framework}(a),
which introduces sufficient spatial-temporal context for representations learning while maintaining a simultaneous manner without absorbing the information of the entire video.

\noindent\textbf{Sequence Encoder.}
The sequence modeling of video segments is based on the stacked multi-head self-attention layers and feed-forward layers,
as proposed in Transformer~\cite{vaswani2017attention} and transformer-based lip reading models~(TM-seq2seq)~\cite{afouras2018deep}.
Moreover, to enable the simultaneous decoding, we employ the time-restricted self-attention, where the unavailable and future 
frames are masked and each video frame can only see its previous several segments $\{{\boldsymbol s}_{t-a+1},{\boldsymbol s}_{t-a+2},...,{\boldsymbol s}_{t}\}$, 
to simulate the streaming inputs and limited history storage. We denote the encoded visual representations of the $t$-th video segment ${\boldsymbol s}_{t}$ as ${\boldsymbol h}_{t}^v$,
as shown in Figure~\ref{fig:framework}.

\subsection{Simultaneous Cross-Modal Decoder}
The simultaneous cross-modal decoder is built based on the neural transducer~\cite{graves2012sequence,jaitly2016online}.
Concretely, at each time step, the decoder~(joint network) chooses to predict the next token ${\boldsymbol w}_{i}$ 
based on the partial input ${\boldsymbol s_{p_{i}}}$, or generate an empty transfer $\epsilon$ 
to read next video segment ${\boldsymbol s}_{n(w_i)+1}$, making $n(w_i) = n(w_i) + 1$.
Also, the syntactic structure of the generated sentences $\{{\boldsymbol w}_{1},{\boldsymbol w}_{2},...,{\boldsymbol w}_{i-1}\}$
are taken into consideration with a language model ${\rm LM(\cdot)}$.
With the reading of video segments, the tokens are generated frame-by-frame and then merged to the ultimate predictions. 

\noindent\textbf{Language Model.} 
Rather than recurrent neural network~(RNN), we also build a uni-directional transformer-based language model that comprises of 
multi-head self-attention and feed-forward layers to leverage the history context of generated sentences.
Specially, the semantic representations of different words are denoted as $\{{\boldsymbol h}_{1}^w,{\boldsymbol h}_{2}^w,...,{\boldsymbol h}_{m}^w\}$.

\noindent\textbf{Joint Network.}
Based on the visual representations given by the simultaneous visual encoder and the semantic representations 
given by the uni-directional language model, we employ a fully-connected layer with softmax to compute the joint matrix ${R}$,
where ${R}_{t,i}$ is the distribution over token vocabulary with ${\boldsymbol h}_{t}^v$ and ${\boldsymbol h}_{w}^i$. 
A possible decoding path ${d} = \{\boldsymbol s_{p_{1}},{\boldsymbol s_{p_{2}}},...,{\boldsymbol s_{p_{u}}}\}$ can be simply
represented as a path from the start $(0,0)$ to the end $(n,m)$ in the joint matrix.
Therefore, the prior optimize object is further denoted as:
\begin{equation}
  {P}_{td}(w|{\boldsymbol s}) = \sum_{{d \in \phi(w)}}{P}(d|{R})
\end{equation}

\subsection{Attention-guided Adaptive Memory}
In real scenarios, the storage is always limited by the extremely long input sequence~(e.g. massive video data).
Therefore, for simultaneous decoding, history segments may be unavailable, making it more difficult to predict a new token with limited visual context.
To achieve a good storage-accuracy trade-off, 
we introduce a novel attention-guided adaptive memory to organize semantic information of history segments and enhance the
visual representations with acceptable computation-aware latency.
Specially, the attention-guided memory containing $k$ memory banks,
is constructed to absorb new segments by momentum update and discard obsolete
features using the least frequently used~(LFU) algorithm guided by attention scores.

\noindent\textbf{Enhanced Visual Feature.}
As shown in Figure~\ref{fig:framework}(b), 
given the adaptive memory ${\boldsymbol M} = \{\boldsymbol m_{{1}},\boldsymbol m_{{2}},...,{\boldsymbol m_{k}}\}$,
we compute a encoder-memory inter-attention for ${\boldsymbol h}_{t}^v$ to enhance the visual representations, given by
\begin{eqnarray}
  \begin{split}
    & {\boldsymbol {\tilde h}}_{t}^v = {\boldsymbol {h}}_{t}^v + \sum_{i=1}^k{\tilde\alpha}_{i}\boldsymbol m_{{i}},~~
    {\tilde\alpha}_{i} = \frac{\exp({\alpha}_{i})}{\sum_{j=1}^k\exp({\alpha}_{j})}
  \end{split}
\end{eqnarray}
where ${\alpha}_{i}$ is the attention score of the $i$-th memory bank $\boldsymbol m_{i}$ and video segment ${\boldsymbol s}_t$,
and ${\boldsymbol {\tilde h}}_{t}^v$ is the enhanced visual feature that absorbs earlier segments.
The enhanced visual feature is actually used for computation of the joint matrix $R$.
Note that we employ the dot-product attention~\cite{vaswani2017attention} to obtain scores over all the memory banks.

\noindent\textbf{Absorb New Segment.}
Since the attention distribution ${\tilde\alpha}$ reflects the similarity between current video segment and
existing segments in the memory banks, some replacements on the memory bank seem to be redundant 
if the segment is close enough to some existing one. To enable higher memory efficiency and avoid storing redundant
information, we adaptively absorb the new segment based on the information entropy ${I_t}$ guided by the attention distributed ${\tilde\alpha}$, given by
\begin{equation}
  {I_t} = -\sum_{i=1}^k{\tilde\alpha_i}\cdot\log({\tilde\alpha_i})
\end{equation}
High information entropy ${I_t}$ represents a more smoothed attention distribution and indicates 
that more information different from the memory is contained in video segment ${\boldsymbol v}_t$,
while low information entropy indicates redundancy. 
To enable higher memory efficiency, we absorb these redundant visual features having ${I_t} < \gamma_e$
by momentum updating,
as shown in Figure~\ref{fig:framework}(b), given by
\begin{equation}
  \begin{split}
    & {\boldsymbol m}_j = {\gamma_m}\cdot {\boldsymbol m}_j + (1-{\gamma_m})\cdot Summarize({\boldsymbol s}_t), \\
    & j = \argmaxA_j{\tilde\alpha_j}
  \end{split}
\end{equation}
where $\gamma_e$ is the information entropy threshold,
${\gamma_m}$ is the parameter to control the impact of moving average, and
$Summarize(\cdot)$ is the operation~(e.g. max-pooling) to aggregate features from different frames within a segment.

\noindent\textbf{Discard Obsolete Segment.}
For these video segments that are distinct from the existing ones in the memory bank, 
we simply replace the least frequently used segment in the adaptive memory.
Also, the counting index is updated based on the soft attention distribution, given by
\begin{equation}
  count({m}_i) = count({m}_i) + \tilde\alpha_i
\end{equation}
And the LFU index is computed as follows:
\begin{equation}
  {LFU}({m}_i) = \frac{count({m}_i)}{life({m}_i)}
\end{equation}
where $count({m}_i)$ and $life({m}_i)$ are separately the counting index of ${\boldsymbol m}_i$ and
timespan that ${\boldsymbol m}_i$ stays in the memory bank. 

\subsection{Training}
\noindent\textbf{Pre-training with CTC Loss.}
To stable the training of the lip reading transducer, we first pre-train the model with an ordinary CTC loss without considering
the syntactic structure of target sequences. 
The CTC also works in frame-synchronized mode and introduces a set of intermediate CTC path $\varphi(w)$ where each path is
composed of target tokens and blanks that can be reduced to the target sequence $w$.
The CTC loss can be computed as follows:
\begin{equation}
  \mathcal{L}_{CTC} = -\log{P_{ctc}(w|s)} = -\log{\sum_{c \in \varphi(w)}P(c|s)}
\end{equation}
With the pre-trained model, we can train the lip reading transducer with the simultaneous lip readig loss as follows:
\begin{equation}
  \mathcal{L}_{SimulLR} = -\log{P}_{td}(w|{\boldsymbol s}) = -\log{\sum_{{d \in \phi(w)}}{P}(d|{R})}
\end{equation}

\noindent\textbf{Model Warm-up.}
Although a better visual encoder~(stacked self-attention and feed-forward layers) can effectively promote the prediction,
it also becomes difficult to train especially with deeper structure for transducer-based methods~\cite{hu2020exploring}.
In this paper, we devise a strategy called model warm-up for the training of lip reading transducer with deeper structure.
Specially, (1) We first apply a shallower sequence encoder~(e.g. less self-attention and feed-forward layers)
and focus on the training of truncated C3D layer, which warms up the C3D encoder.
(2) We then freeze the parameters of truncated C3D and employ a deeper network structure, which warms up the sequence encoder.
(3) We train both the visual encoder that has been warmed up
and simultaneous decoder with the proposed loss.

\noindent\textbf{Curriculum Learning.}
To further make the training procedure stable, we exploit the novel training paradigm based on the curriculum learning
that starts with short videos, learns the easier aspects of lip reading and then gradually increase the length of
training videos.

\section{Experiments}
\subsection{Datasets}
\noindent\textbf{GRID.} 
The GRID~\cite{cooke2006audio} dataset contains 34,000 sentences uttered by 34 speakers.
This dataset is easy to learn since the spoken sentences are in a restricted grammar and composed of 6$\sim$10 words.
The vocabulary of GRID is also small, comprising 51 different words including 4 commands, 4 color, 4 prepositions,
25 letters, 10 digits and 4 adverbs.
All the videos of lip movements have the same length of 75 frames with a frame rate of 25fps.
Following prior works~\cite{assael2016lipnet,chen2020duallip,liu2020fastlr}, we randomly select 255 sentences 
for evaluation.

\noindent\textbf{TCD-TIMIT.} 
The TCD-TIMIT~\cite{harte2015tcd} dataset contains 59 speakers that utters approximately 100 phonetically rich sentences,
making this dataset more challenging but closer to the natural scene.
Also, the video length and sentence in the TCD-TIMIT dataset are longer than GRID and variable.
Following prior work~\cite{harte2015tcd}, we use the recommended train-test splits for training and evaluation.

\subsection{Implementation Details}
\noindent\textbf{Data Preprocessing.}
For the videos, to extract lip movements, we first obtain a $256 \times 256$ aligned face with Dlib detector~\cite{king2009dlib}, 
crop the $160 \times 80$ mouth-centered region from the aligned face and then resize the region to $100 \times 60$ as the video input.
To improve the recognition accuracy, we use the strategy of data augmentation that involves horizontal flips with 40\% probability,
crop 0$\sim$5\% of horizontal or vertical pixels with 40\% probability.
In particular, we convert the video frames to grey scale for the easier GRID dataset to reduce computation cost.
For the sentences, we build a vocabulary at word-level for the GRID dataset while phoneme-level for the TCD-TIMIT dataset
following previous works~\cite{harte2015tcd}.

\noindent\textbf{Model Setting.}
For simultaneous decoding, we set the number of available segments $a$ to 2, and the number of frames in a video segment
$n_f$ to 3 for GRID dataset and to 5 for TCD-TIMIT dataset. 
The number of memory banks $k$ is set to 20.
The information entropy threshold $\gamma_e$ is set to $0.6 \times \log_2{k}$ and moving step $\gamma_m$ is set to 0.7.
For the truncated C3D to extract spatial-temporal representations, we stack six 3D convolutional layers with 3D max pooling,
RELU activation, and two fully connected layers. The kernel size of 3D convolution and pooling is set to $3 \times 3$.
For both the segment sequence encoder and language model, we stack four self-attention layers with feed-forward network.
We set $d_{hidden} = 256$ for GRID dataset and $d_{hidden} = 512$ for TCD-TIMIT dataset respectively.
The joint network is simply a two-layer non-linear transformation.

\noindent\textbf{Training Setup.}
For GRID dataset, We pretrain the model using the CTC loss with 10 epochs, warmup the visual encoder using two sequence encoder layers
with 20 epochs and then train the whole model using four encoder layers with 100 epochs.
For TCD-TIMIT dataset, We pretrain the model using the CTC loss with 50 epochs, warmup the visual encoder using two sequence encoder layers
with 50 epochs and then train the whole model using four encoder layers with 150 epochs.
To train the SimulLR model, we employ the Adam optimizer with a initial learning rate 0.0005 for GRID dataset and 0.0003 for TCD-TIMIT dataset,
and with a shrink rate of 0.99 according to the updating step.

\subsection{Evaluation Metrics}
During the inference stage, the SimulLR model perform simultaneous decoding with the adaptive memory.
Following prior works~\cite{chen2020duallip}, to evaluate the recognition quality, we use the metrics of 
character error rate~(CER) and word error rate~(WER) on the GRID dataset, and phoneme error rate~(PER) 
on the TCD-TIMIT dataset since the output of this dataset is phoneme sequence. 
The different types of error rate can be computed as follows:
\begin{equation}
  ErrorRate = \frac{(S + D + I)}{M}
\end{equation}
where $S, D, I, M$ are separately the number of the substitutios, deletions, insertions and reference
tokens~(character, word or phoneme).

To compute the latency of simultaneous decoding, we consider the non computation-aware~(NCA) latency, 
as proposed in~\cite{ma2020simulmt}.
Specially,
The NCA latency for ${\boldsymbol w}_i$, $d_{NCA}({w}_i)$, equals to $n({w}_i)\cdot n_f\cdot{T_s}$,
where $T_s$ (ms) is the frame sampling rate.
The average NCA latency $AL_{NCA}$ is defined as:
\begin{equation}
  AL_{NCA} = \frac{1}{\tau({w})}\sum_{i=1}^{\tau(w)}d_{NCA}({w}_i) - r\cdot(i - 1)\cdot{T_s}
\end{equation}
where $\tau({w})$ denotes the index of the first generated token when the model read the entire video,
and $r = {(n \cdot n_f)}/{u}$ is the length ratio between source and target sequence. 

\begin{table}[t]
  \centering
  \caption{The word error rate~(WER) and character error rate~(CER) on the GRID dataset, and
  the phoneme error rate~(PER) on the TCD-TIMIT dataset.}
  \scalebox{1.0}{
  \begin{tabular}{ccccc}
      \toprule
      \multicolumn{2}{l}{}       & \multicolumn{2}{c}{GRID} & TCD-TIMIT \\
      \cmidrule(lr){3-4} \cmidrule(lr){5-5}
      \multicolumn{2}{l}{Method} & WER(\%) & CER(\%) & PER(\%)\\
      \midrule
      \multicolumn{4}{l}{\textit{Non-Simultaneous Methods}} \\ 
      \midrule
      \multicolumn{2}{l}{LSTM~\cite{wand2016lipreading}}      & 20.4    & /   & / \\
      \multicolumn{2}{l}{LipNet~\cite{assael2016lipnet}}      & 4.8     & 1.9 & /\\
      \multicolumn{2}{l}{FastLR~\cite{liu2020fastlr}}         & 4.5     & 2.4 & /\\
      \multicolumn{2}{l}{LCANet~\cite{xu2018lcanet}}          & 4.215   & 1.532 & /\\
      \multicolumn{2}{l}{DualLip~\cite{chen2020duallip}}      & {\bf 2.71}    & {\bf 1.16} & {\bf 46.2} \\
      \midrule
      \multicolumn{4}{l}{\textit{Simultaneous Methods}} \\
      \midrule
      \multicolumn{2}{l}{LR-RNN-CTC}              & 28.884 & 19.912 & 67.021\\
      \multicolumn{2}{l}{LR-TM-CTC}      & 20.691 & 15.223 & 63.428 \\
      \multicolumn{2}{l}{LR-RNN-TD}       & 11.570 & 7.263 & 64.213\\
      \multicolumn{2}{l}{LR-TM-TD}      & {3.125} & {1.588}  & {62.831} \\
      \multicolumn{2}{l}{SimulLR(Ours)} & {\bf 2.738} & {\bf 1.201} & {\bf 56.029}\\
      \bottomrule
  \end{tabular}
}
\label{table:main_result}
\end{table}

\begin{table}[t]
  \centering
  \caption{The comparison of NCA latency and corresponding recognition accuracy 
  with different segment size $n_f$ on TCD-TIMIT dataset.
  The evaluation is conducted with 1 Nvidia 2080Ti GPU. }
  \scalebox{1.0}{
  \begin{tabular}{ccccc}
      \toprule
      \multicolumn{2}{l}{Methods}  & PER(\%) & Latency(ms) & Speedup\\
      \midrule
      \multicolumn{2}{l}{DualLip~\cite{chen2020duallip}}                 & 46.200   & 4580.0 & 1.00$\times$\\
      \multicolumn{2}{l}{SimulLR~($n_f = 3$)}     & {58.182} & 384.93  & 11.91$\times$\\
      \multicolumn{2}{l}{SimulLR~($n_f = 5$)}     & {56.029} & 502.83  & 9.10$\times$\\
      \multicolumn{2}{l}{SimulLR~($n_f = 20$)}    & {49.743} & 973.62  & 4.70$\times$\\
      \bottomrule
  \end{tabular}
}
\label{table:latency_result}
\end{table}

\subsection{Main Results}
Since prior methods are trained in a non-simultaneous setting, to verify the effectiveness of our proposed methods,
we first build several simultaneous lip reading baselines as follows:

\noindent\textbf{LR-RNN-CTC.} 
Using the convolutional network and uni-directional recurrent neural network as visual encoder, we train the simultaneous
model with the mentioned CTC loss.
Note that RNN is already a natural memory network to organize the history information.

\noindent\textbf{LR-RNN-TD.} 
Further considering the syntactic structure of generated sequences, we introduces language model and train the simultaneous
model with the transducer loss.

\noindent\textbf{LR-TM-CTC.}
By replacing the RNN sequence encoder with the popular transformer architecture, we train the model with the mentioned CTC loss.

\noindent\textbf{LR-TM-TD.}
With the transformer architecture, we introduce the language model and train the network with the transducer loss.

We compare our methods with some mainstream state-of-the-art non-simultaneous models and the constructed baselines.
The overall evaluation results on two datasets are presents in Table~\ref{table:main_result}.
We can see that (1) The proposed SimulLR outperforms all the simultaneous baselines by a large margin, indicating
the effectiveness of our method for simultaneous lip reading. (2) The SimulLR also achieves comparable results 
with the state-of-the-art non-simultaneous method DualLip~\cite{chen2020duallip}, especially on GRID datasets,
demonstrating the potential of our method. (3) With the same visual encoder, the transducer-based models obtain better
performance than CTC-based models, verifying the effectiveness of modeling of syntactic structure.

\begin{figure}[t]
  \includegraphics[width=0.9\linewidth]{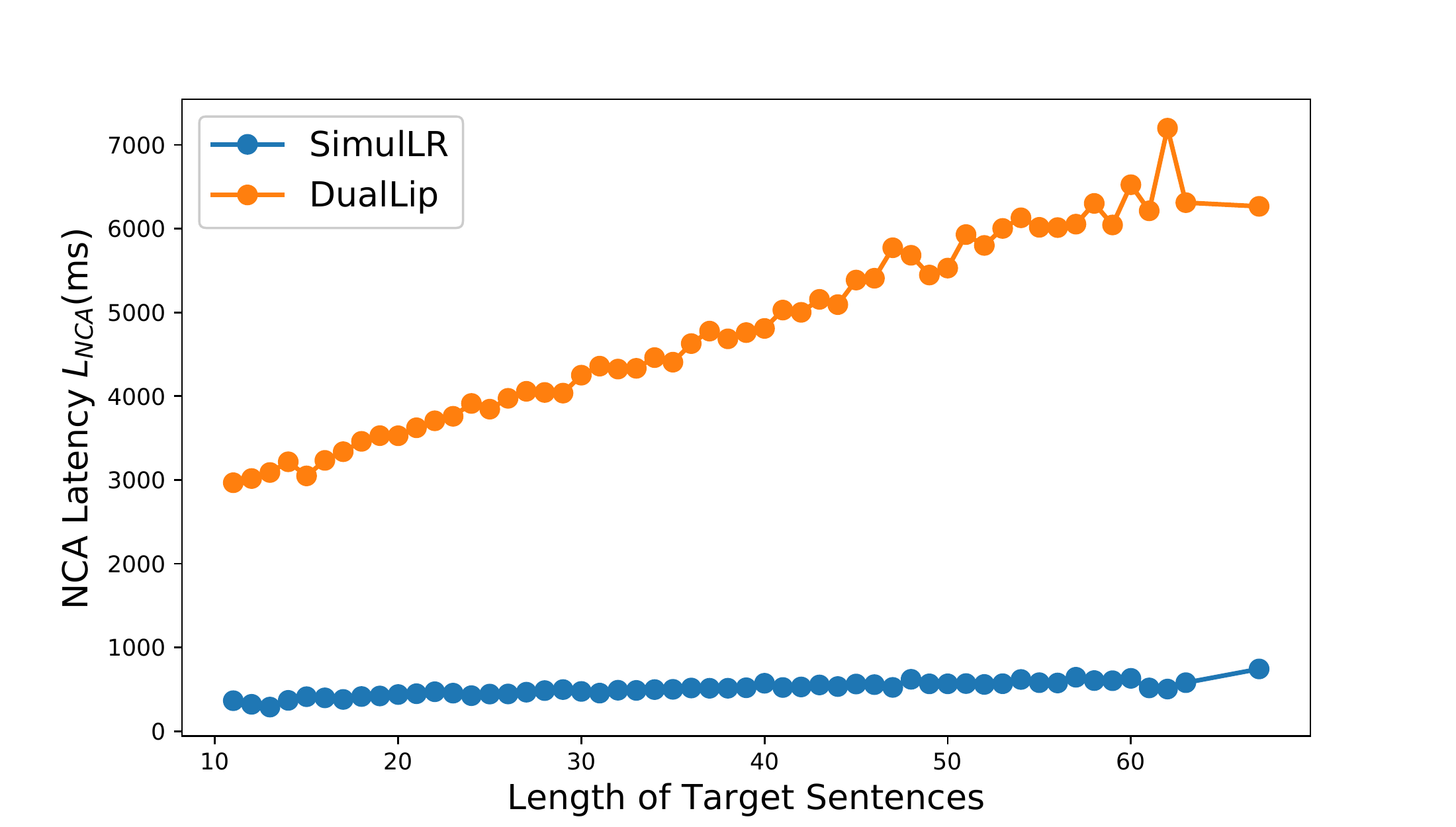}
  \caption{
      The NCA latency of target sentences with different length for DualLip and SimulLR
      on TCD-TIMIT dataset. 
  }
  \label{fig:nca_length}
\end{figure}
 
\begin{figure}[t]
  \includegraphics[width=0.9\linewidth]{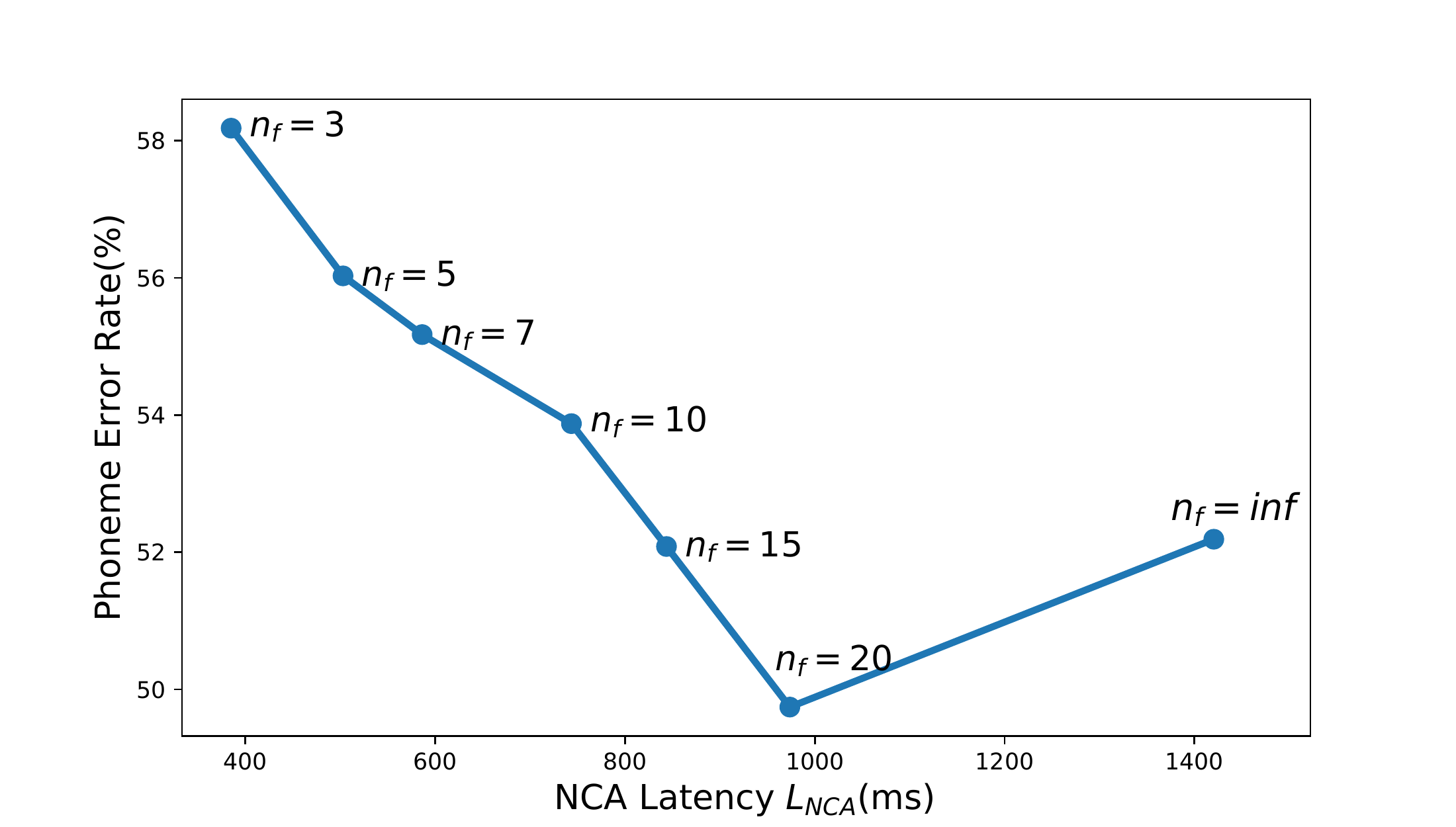}
  \caption{
      The recognition accuracy against the NCA latency with different segment size $n_f$
      on TCD-TIMIT dataset. 
  }
  \label{fig:nca_tcd}
\end{figure}

\subsection{Latency Analysis}
In this section, to further explore the time-efficiency of the proposed SimulLR method, 
we record the prediction latency of both simultaneous and non-simultaneous models
to make a comparison.

We first measure the inference NCA latency and corresponding recognition accuracy of DualLip~\cite{chen2020duallip} and SimulLR
with $n_f = 5$ and $n_f = 20$, which is listed in Table~\ref{table:latency_result}.
As the results shows, compared with the non-simultaneous method DualLip,
the SimulLR speeds up the prediction by 9.10$\times$ with $n_f = 5$, and 4.70$\times$ with $n_f = 20$.
Also, the SimulLR~($n_f = 20$) even achieves competitive results~(PER 49.743\%) with less waiting time,
indicating the great ability of the adaptive memory to incorporate history information.
The speedup rate increases rapidly especially for longer sentences, as shown in Figure~\ref{fig:nca_length}.
During inference, the non-simultaneous models wait for the entire video to process, making the NCA latency increases
with respect to the length of target sequence, while the NCA latency nearly holds a small constant for SimulLR.
Further, considering the computation-aware~(CA) latency 
that is the time elapsing from the processing of corresponding input to the prediction of a token,
compared with attention-based TM-Seq2Seq~\cite{afouras2018deep},
the SimulLR achieves a speedup of 1.6$\times$ on GPU and 13.3$\times$ on CPU, indicating the effectiveness 
of memory to reduce the computation-aware cost.

To explore the performance of simultaneous decoding, we also measure the NCA latency and phoneme error rate with different segment size $n_f$
on TCD-TIMIT dataset, as shown in Figure~\ref{fig:nca_tcd}.
Note that for ``$n_f = {\rm inf}$'', we remove the memory and all the history segments are available.
The recognition accuracy increases as the segment size increases, with the sacrifice of NCA latency.
Notice that the SimulLR~($n_f = 20$) even obtain better performance than model with all the history segments.
which indicates that compared with direct interaction with all the history segments,
the proposed memory can better organize history information, discard obsolete segments
and extract useful context for prediction.

\begin{table}[t]
  \centering
  \caption{The ablation results on the GRID and TCD-TIMIT dataset.
  We add the proposed techniques and evaluate their effectiveness progressively.}
  \scalebox{0.9}{
  \begin{tabular}{ccccc}
      
      \toprule
      \multicolumn{2}{l}{}       & \multicolumn{2}{c}{GRID} & TCD-TIMIT \\
      \cmidrule(lr){3-4} \cmidrule(lr){5-5}
      \multicolumn{2}{l}{Models} & WER(\%) & CER(\%) & PER(\%)\\
      \midrule
      \multicolumn{2}{l}{Naive LR Transducer}      & {3.125} & {1.588}  & {62.831}\\
      \midrule
      \multicolumn{2}{l}{+CTC}      & {3.029} & {1.503} & {62.032}\\
      \multicolumn{2}{l}{+CTC+TC3D}      & {2.978} & {1.339} & {60.397}\\
      \multicolumn{2}{l}{+CTC+TC3D+WARM}   & {2.963}  & {1.302} & {59.428}\\
      \midrule
      \multicolumn{2}{l}{+CTC+TC3D+WARM+MEM~}      \\
      \multicolumn{2}{l}{(SimulLR)}      & {\bf 2.738} & {\bf 1.201} & {\bf 56.029}\\
      \bottomrule
  \end{tabular}
}
\label{table:ab_result}
\end{table}

\begin{table}[t]
  \centering
  \caption{The effect of different memory strategies on the GRID and TCD-TIMIT dataset.}
  \scalebox{1.0}{
  \begin{tabular}{ccccc}
      
      \toprule
      \multicolumn{2}{l}{}       & \multicolumn{2}{c}{GRID} & TCD-TIMIT \\
      \cmidrule(lr){3-4} \cmidrule(lr){5-5}
      \multicolumn{2}{l}{Memory strategy} & WER(\%) & CER(\%) & PER(\%)\\
      \midrule
      \multicolumn{2}{l}{FIFO Queue}    & 2.894 & 1.313  & 57.731\\
      \multicolumn{2}{l}{LFU}           & 2.881 & 1.292 & 57.384\\
      \multicolumn{2}{l}{Ours~(LFU + Momentum)}     & {\bf 2.738} & {\bf 1.201} & {\bf 56.029}\\
      \bottomrule
  \end{tabular}
}
\label{table:mem_type_result}
\end{table}

\subsection{Ablation Analysis}
In this section, to explore the effectiveness of the proposed techniques in SimulLR, we first conduct ablation experiments
on the GRID and TCD-TIMIT datasets. 
The evaluation results are presented in Table~\ref{table:ab_result}.

\noindent\textbf{Naive LR Transducer~(Base).} 
We construct the base model with only the convolutional network and transformer architecture as visual encoder
and the frame-synchronized simultaneous transducer-based decoder.

\noindent\textbf{Base+CTC.}
To stable the training and promote the performance, we first employ the CTC pre-training for base model, and
the results demonstrate that CTC pre-training is helpful for the cross-modal alignment between visual frames 
and textual tokens.

\noindent\textbf{Base+CTC+TC3D.}
To enhance the visual representations while maintaining the simultaneous manner, we replace the 2D convolutional network
with truncated C3D layer in the visual encoder, and the results show that the truncated C3D layer can effectively
boost the feature representation ability of visual encoder.

\noindent\textbf{Base+CTC+TC3D+WARM.}
To further improve the performance, we apply the proposed model warm-up strategy where we
train a deeper network step by step.
As shown by the results, the warm-up technique can further facilitate the features learning of visual encoder and
improve the performance.

\noindent\textbf{Base+CTC+TC3D+WARM+MEM.}
With limited history and to reduce the computational cost, we further add the proposed attention-guided adaptive memory
to organize semantic information of history segments and enhance the visual representations.
Table~\ref{table:ab_result} shows that using the adaptive memory can boost the performance significantly, showing that
the proposed memory can effectively organize history information and incorporate global context for visual representations
enhancement.

\begin{table}[t]
  \centering
  \caption{The effect of memory size $k$ on the GRID and TCD-TIMIT dataset.}
  \scalebox{1.0}{
  \begin{tabular}{ccccc}
      
      \toprule
      \multicolumn{2}{l}{}       & \multicolumn{2}{c}{GRID} & TCD-TIMIT \\
      \cmidrule(lr){3-4} \cmidrule(lr){5-5}
      \multicolumn{2}{l}{Memory size} & WER(\%) & CER(\%) & PER(\%)\\
      \midrule
      \multicolumn{2}{l}{k = 5}       & 2.881 & 1.289  & 56.896\\
      \multicolumn{2}{l}{k = 10}      & 2.796 & 1.259 & 56.528\\
      \multicolumn{2}{l}{k = 20}      & {\bf 2.738} & {\bf 1.201} & {\bf 56.029}\\
      \multicolumn{2}{l}{k = inf}     & {2.760} & {1.236} & {56.173}\\
      \bottomrule
  \end{tabular}
}
\label{table:mem_size_result}
\end{table}

\begin{table}[t]
  \centering
  \caption{The effect of different ways of summarization on the GRID and TCD-TIMIT dataset.}
  \scalebox{1.0}{
  \begin{tabular}{ccccc}
      
      \toprule
      \multicolumn{2}{l}{}       & \multicolumn{2}{c}{GRID} & TCD-TIMIT \\
      \cmidrule(lr){3-4} \cmidrule(lr){5-5}
      \multicolumn{2}{l}{Summarization} & WER(\%) & CER(\%) & PER(\%)\\
      \midrule
      \multicolumn{2}{l}{conv}      & {2.872} & {1.286} & {57.461}\\
      \multicolumn{2}{l}{max-pooling}      & {2.763} & {1.267} & {57.116}\\
      \multicolumn{2}{l}{avg-pooling}      & {\bf 2.738} & {\bf 1.201} & {\bf 56.029}\\
      \bottomrule
  \end{tabular}
}
\label{table:summarize_result}
\end{table}

Besides, we further study the effectiveness of the proposed adaptive memory from the perspective of
the memory strategy, the memory size and the way to summarize semantic information from new segments.
As results shows in Table~\ref{table:mem_type_result}, we first devise different memory strategies to organize history segments 
including first-in-first-out~(FIFO) queue, and attention-guided least frequently used~(LFU) algorithm.
The memory with FIFO queue achieves the worst performance, demonstrating that the LFU can effectively extract useful
history information, while the adaptive memory with momentum updating obtains the best performance, indicating that
the attention-guided strategy with entropy can avoid storing redundant information and enable higher memory efficiency.

We then explore the effect of memory size on the recognition accuracy. 
As shown in Table~\ref{table:mem_size_result}, increasing the memory size $k$ can firstly absorb more history information
for better recognition, while there is an error rate increase for ``$k = {\rm inf}$'' where relatively unimportant contexts are introduced
as noise.

We also conduct different ways to summarize semantic information from new segments including convolution, max-pooling
and avg-pooling, and the evaluation results are presented in Table~\ref{table:summarize_result}.
Compared with ``conv'' and ``max-pooling'', the ``avg-pooling'' can better summarize the semantic information of a video
segment.

\begin{figure}[t]
  \includegraphics[width=0.9\linewidth]{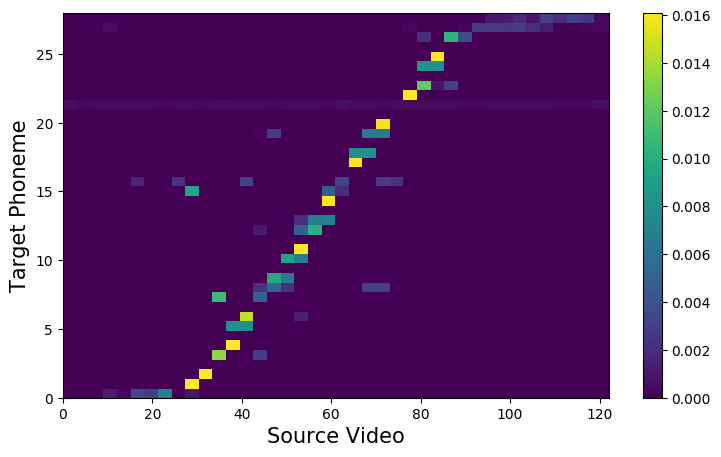}
  \includegraphics[width=0.9\linewidth]{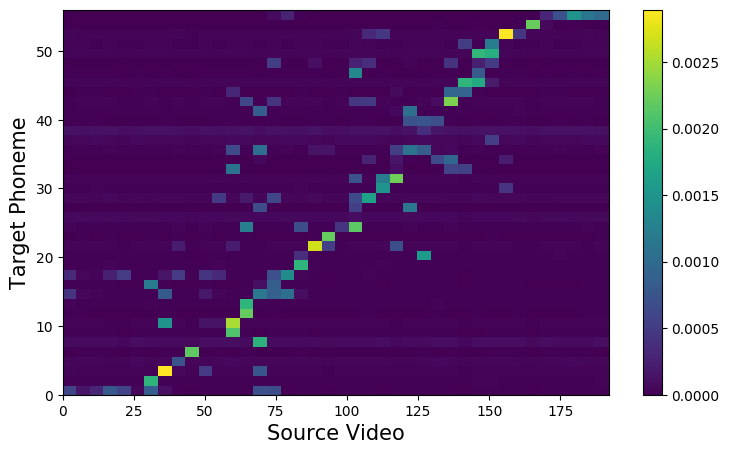}
  \caption{
    The visualization of monotonic alignment between target sequence and source video on TCD-TIMIT dataset~(memory size $k = 10$).
    The second row represents a longer video with more frames.
    The brightness of the color represents the degree of alignment between tokens and frames.
  }
  \label{fig:tcd_ali}
\end{figure}

\subsection{Qualitative Results}
As shown in Figure~\ref{fig:tcd_ali}, by normalizing the distribution of target token over all frames,
we visualize the monotonic alignment learned by SimulLR between target sequence and source video on TCD-TIMIT dataset.
The brightness of the color represents the matching degree between tokens and frames.
The approximate monotonic alignment in the figure indicates the effectiveness of our proposed methods to learn the cross-modal alignment under simultaneous setting
and with limited history.


  

\section{Conclusions}
In this paper, we study the task of simultaneous lip reading and devise SimulLR,
a simultaneous lip Reading transducer with attention-guided adaptive memory.
To address the challenging of monotonic alignments while considering the syntactic structure of the generated sentences,
we build a transducer-based model with adaptive memory and design several effective training strategies including CTC pre-training, model warm-up and curriculum learning to promote the training
of the lip reading transducer.
Also, to learn better spatio-temporal representations for simultaneous encoder, we construct a truncated 3D convolution and time-restricted
self-attention layer to perform the frame-to-frame interaction within a video segment.
Further, the history information is always limited due to the storage in real-time scenarios.
To achieve a good trade-off, we devise a novel attention-guided adaptive memory to organize semantic information of history segments and enhance the
visual representations with acceptable computation-aware latency.
Experiments on GRID and TCD-TIMIT datasets shows that the SimulLR outperforms the baselines and has great time-efficiency,
demonstrating the effectiveness of our methods for simultaneous lip reading.

\begin{acks}
  This work was supported in part by the National Key R\&D Program of China under Grant No.2018AAA0100603, 
  National Natural Science Foundation of China under Grant No.61836002, No.62072397 
  and Zhejiang Natural Science Foundation under Grant LR19F020006.
\end{acks}

\bibliographystyle{ACM-Reference-Format}
\balance
\bibliography{ref.bib}

\end{document}